\documentclass[accepted]{uai2024} % after acceptance, for a revised version; 
\usepackage[british]{babel}

\usepackage{mathtools} % amsmath with fixes and additions
\usepackage{booktabs} % commands to create good-looking tables
\usepackage{tikz} % nice language for creating drawings and diagrams

\usepackage{tikz}
\usepackage{soul}
\usepackage{xcolor}
\usepackage{mathtools}
\usepackage{adjustbox}
\usepackage{tabularx}
\usepackage{adjustbox}
\usepackage{siunitx}
\usepackage{subcaption}
\usepackage{bm}
\usepackage{upgreek}
\usepackage[square,sort,comma,authoryear,round]{natbib}
\usepackage{hyperref}
\hypersetup{colorlinks,allcolors=black}
\usepackage{multirow}
\usepackage{makecell}
\usepackage{graphicx}
\usepackage{amssymb}
\usepackage{amsthm}
\usepackage{booktabs}
\usepackage{amsmath}
\usepackage[utf8]{inputenc}
\usepackage{varioref}
\usepackage{cleveref}
\usepackage{graphicx}
\usepackage{enumitem}
\usepackage{amsmath}
\usepackage{multirow}
\usepackage{tcolorbox}
\usepackage{xcolor}
\usepackage{bbm}
\usepackage{authblk}
\usepackage{adjustbox}
\usepackage{appendix}
\usepackage{url}
\usepackage{algorithm} % For algorithm
\usepackage{algpseudocode} % For algorithm
\usepackage{arydshln} % hdashline

\graphicspath{{images/}}
\usepackage{makecell, booktabs}

\renewcommand{\vec}[1]{\mathbf{#1}}

\renewcommand{\vec}[1]{\mathbf{#1}}

\usepackage{natbib} % has a nice set of citation styles and commands
\graphicspath{{images/}}
\newcommand{\lbl}{\ell}
\renewcommand{\vec}[1]{\mathbf{#1}}
\newcommand{\dpool}{$\mathcal{D}^{\text {pool }}$}
\newcommand{\dtrain}{$\mathcal{D}^{\text {train }}$}
\newcommand{\dtest}{$\mathcal{D}^{\text {test }}$}
\newcommand{\dval}{$\mathcal{D}^{\text {val }}$}
\newcommand{\sbald}{${\text {BALD}}$}
\newcommand{\scbald}{{$\mathcal{C}\text{-BALD }$}}
\newcommand{\srandom}{$\text {Random }$}
\newcommand{\sunc}{${\text {Entropy}}$}

\usepackage{xpatch}
\usepackage{xcolor}
\usepackage{booktabs}
\usepackage{tabularx}
\usepackage{makecell}
\usepackage{amsthm}
\usepackage{natbib}
\usepackage{amssymb}
\usepackage{tikz}
\usetikzlibrary{bayesnet}
\definecolor{myred}{rgb}{0.8594, 0.2695, 0.1914}
\definecolor{myblue}{rgb}{0.122, 0.467, 0.706}
\definecolor{myorange}{rgb}{1,0.49,0.05}
\xpretocmd{\eqref}{Equation~}{}{}

\title{Bayesian Active Learning for Censored Regression}

\author[1]{Frederik Boe Hüttel}
\author[1]{Christoffer Riis}
\author[1]{Filipe Rodrigues}
\author[1]{Francisco C. Pereira}
\affil[1]{%
Technical University of Denmark%,  Denmark
}
  
\begin{document}
\maketitle

\begin{abstract}
Bayesian active learning is based on information theoretical approaches that focus on maximising the information that new observations provide to the model parameters.
This is commonly done by maximising the Bayesian Active Learning by Disagreement (BALD) acquisitions function.
However, we highlight that it is challenging to estimate BALD when the new data points are subject to censorship, where only clipped values of the targets are observed.
To address this, we derive the entropy and the mutual information for censored distributions and derive the BALD objective for active learning in censored regression ($\mathcal{C}$-BALD).
We propose a novel modelling approach to estimate the $\mathcal{C}$-BALD objective and use it for active learning in the censored setting.
Across a wide range of datasets and models, we demonstrate that $\mathcal{C}$-BALD outperforms other Bayesian active learning methods in censored regression.
\end{abstract}

\section{Introduction}
\begin{figure}[t]
    \includegraphics[width=\columnwidth]{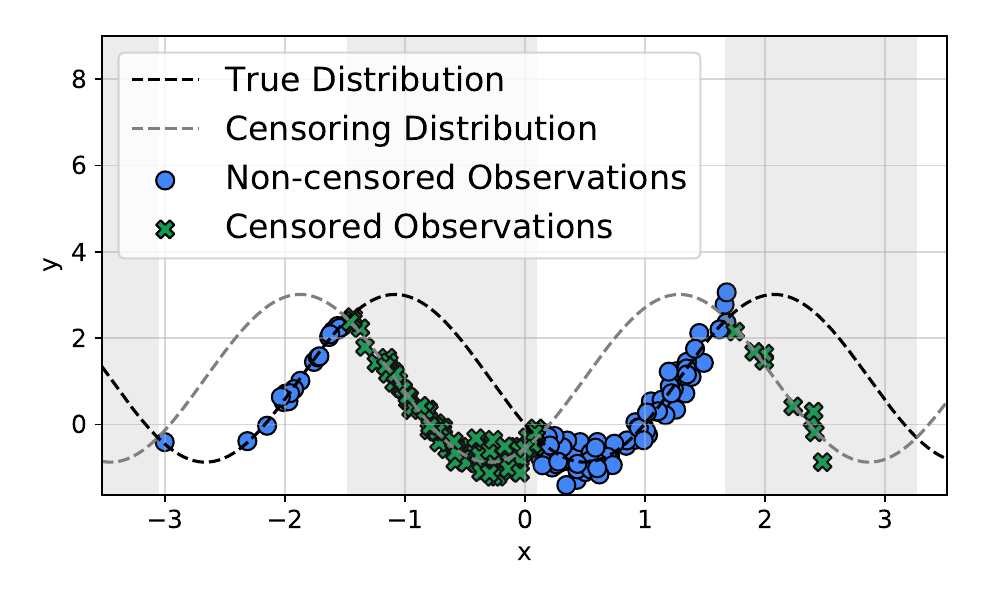}
    \caption{Illustration of a 1-dimensional censored dataset, in which the dashed black line represents the underlying function that generated the data. The (blue) circles denote non-censored observations, while the (green) crosses represent observations that have been censored.
The grey background indicates areas where the observations are censored.}
    \label{fig:dataexample}
\end{figure}

Active learning is a framework where a model learns from a small amount of labelled data and chooses the data it wants to acquire a label for \citep{Settles2009}. 
This acquisition of new data points is done iteratively to improve the model's predictive performance and reduce model uncertainty \citep{mackay1992theevidence}.
This naturally poses the challenge: which new data points can improve the model the most?
Information theoretical approaches are often the basis to solve this challenge by reasoning about the information that new labels can provide to the model's parameters \citep{mackay1992information}. 
A common method to estimate the information is with the \emph{Bayesian Active Learning by Disagreement} (BALD) acquisition function~\citep{houlsby2011bayesian}, which estimates the mutual information between the model parameters and the labels to identify which labels to acquire.
It has successfully been applied to various domains, such as computer vision \citep{gal2017deep, kirsch2019batchbald} and natural language processing~\citep{shen2018deep}.

Even though existing Bayesian active learning methods have proven useful, using them for censored regression problems remains challenging.
The challenge in censored regression is that the target variable is only partially observed \citep{tobin1958estimation, cox1972regression, power1986censored}.
To illustrate the concept, consider the dataset shown in Figure \ref{fig:dataexample}. The data points in this set are generated from a true (\emph{latent}) function, where some targets are censored, meaning that only clipped values of the true function are observed. The green crosses in the figure represent censored observations whose $y$-values, as a consequence of the censoring, are distributed below the mean of the true function.
In contrast, the blue circles in the figure are non-censored observations and follow the true generating function.

Censored regression models try to approximate the true function using \emph{both} censored and non-censored observations~\citep{huttel2022modelling}.  
Naturally, censored data presents unique modelling challenges to determine which new data points provide the most information, as censored observations do not provide the same information about the true function as uncensored ones~\citep{hollander1985information, hollander1987measuring, hollander1990information}.
As a result, the BALD objective can not directly be applied to censored regression problems because it measures the information gained by observing the true function, where, in reality, we might observe a clipped value.

This issue has considerable practical implications, for example, in shared mobility services, such as shared bikes, shared electric vehicles or electric vehicle charging, where the supply censors the observed demand~\citep{gammelli2020estimating, huttel2023mind}.
Modern machine learning models are becoming more integral to managing these systems. 
This raises the question \citep{golsefidi2023ajoint}: where should operators focus their attention to gather the most informative data for their model in a city? 
The same challenges can be found beyond the transport domain, such as in 
subscription-based businesses (e.g. what customers will cancel their subscription) \citep{FADER200776, chandar2022usingsurv, maystre2022temporallyconsistent}, and in health survival applications where labels are expensive to collect (e.g. in medical imagining or invasive diagnosing) \citep{zafar2019adeepactive, lain2022imageing}.

Motivated by these challenges, we study Bayesian active learning for censored regression problems, focusing on estimating the information new observations provide to the model parameters. 
This is challenging because new labels might be subject to censorship and, therefore, do not have the potential to provide the same amount of information as uncensored ones.
%We base our approach on theoretical information principles to reason about the information new labels provide to the model.
Concretely, this paper makes the following contributions:
\begin{enumerate}
    \item Formulation and derivation of the mutual information between observations and model parameters when the observations are subject to censoring. 
    \item A novel acquisition function for active learning using the derived mutual information and the entropy of censored distributions and a novel modelling approach to estimate the entropy.
    %\item A novel modelling approach to estimate the entropy and mutual information, 
    \item Evaluation of the proposed acquisitions on synthetic and real-world datasets compared to Bayesian active learning methods.
\end{enumerate}

\section{Background \& Setting}
We are interested in the supervised learning of a probabilistic regression model, $p(y^*_i| \vec{x}_i, \theta)$, where $\vec{x}_i \in \mathcal{X} \subseteq \mathbb{R}^d$ for $d \geq 1$, $y^*_i \in \mathcal{Y}^* \subseteq \mathbb{R}$, and $\theta$ is a set of stochastic model parameters.
We assume that we can sample a set of model parameters, $\theta$, from the posterior distribution $p(\theta|\mathcal{D})$.
We consider the special regression case, where $y^*_i$ is subject to censoring, meaning that for some observations in our dataset, $y^*_i$ is unknown.
Specifically, we consider \emph{right-censored} data, which means that instead of observing $y^*$, we observe $y_i= \min(y^*_i, z_i)$, where $z_i \in \mathcal{Z} \subseteq \mathbb{R}$.
$z_i$ is a censoring threshold.
In addition, we also observe a censoring indicator $\lbl_i=\mathbbm{1}\{y^*_i \leq z_i\}$, which indicates whether $y_i$ is censored or not.
%This data type is called \emph{right-censored} data. 
A censored dataset of size $n$ can thus be denoted $\mathcal{D}=\{\vec{x}_i, y_i, \lbl_i\}_{i=1}^n$.
For readability and to simplify the notation, we will use $p(x)$ to denote the density function of a random variable $x$.

In the case of censored regression, the objective is to infer the true distribution $p(y_i^*| x_i, \theta)$ and the model parameters, $\theta$, based on the censored dataset $\mathcal{D}$. 
%This is challenging because $y_i^*$ is only partially observed.
In censored regression, one typically assumes that the distributions of $p(y_i^*| \vec{x}_i)$ and $p(z_i| \vec{x}_i)$ are independent given the covariates, $\vec{x}_i$ \citep{tobin1958estimation}. 
This assumption is more general than other assumptions, such as fixed-value censoring, i.e., $z_i=\text{constant } \forall i$~\citep{power1986censored}.
We formally state the following assumption.

\textbf{Assumption 1.} \textit{(Independent censoring) We assume that conditioned on the covariates, $\vec{x}_i$, the censoring distribution and the true distribution of the target are independent. That is, we assume that $y_i^* \perp z_i | \vec{x}_i$}.

Under Assumption 1, the right censored log-likelihood is defined as,
\begin{equation}
\label{eq:tobit_fit}
\begin{aligned}
\mathcal{L}_\text{C}\left(\theta\right)= -\sum_{i \in \mathcal{D}}  &\Big( \underbrace{\lbl_i\log \left(\varphi\left(y_i | \vec{x}_i, \theta \right) \right)}_{\text{Observed loss}} +\\
& \underbrace{(1-\lbl_i)\log\left(1-\Phi\left(z_i| \vec{x}_i, \theta \right)\right)}_{\text{Censored loss}} \Big)\, ,
\end{aligned}
\end{equation}
where $\Phi$ is the Cumulative Distribution Function (CDF) and $\varphi$ is the Probability Density Function (PDF) of $p(y^*_i| \vec{x}_i, \theta)$.
The term $(1-\Phi)$ is often called the censoring or survival distribution.
While we focus on right-censoring, left censoring (i.e. $y_i=\max(y^*_i,z_i)$) can be handled by inverting $y_i \forall i$.

The class of models that can fit the distribution for $\varphi$ and $\Phi$  is broad. 
It includes, in practice, all Bayesian models for which $p(y_i^* | x_i, \theta)$ has a fixed PDF and CDF, and $p(\theta)=p(\theta|\mathcal{D})$ is the posterior distribution given the observed dataset $\mathcal{D}$. Common models include deep ensembles~\citep{lakshminarayanan2017simple} and neural networks with stochastic parameters \citep{gal16dropout, sharma2023dobayesian}.

\subsection{Active Learning}
In the supervised setting, active learning involves having a model select which labels to acquire during training to increase the model performance \citep{mackay1992information, Settles2009}.
It maximises an acquisition function, which captures the utility of acquiring the label for a given input~\citep{kirsch2022unifying}.
We are interested in such settings, but where the data points are subject to censoring.

Typically, one starts with a small training dataset, 
\begin{equation}
\mathcal{D}^{\text {train }}=\left\{\left(\vec{x}_i, y_i, \lbl_i\right)\right\}_{i=1}^n \, .
\end{equation}
which is used to train a probabilistic model with likelihood $p(y_i^* | \vec{x}_i, \theta)$.
Then, from a larger (finite or infinite) pool of unlabelled data,
\begin{equation}
\mathcal{D}^{\text {pool }}=\left\{\vec{x}_i\right\}_{t=1}^m,
\end{equation}
the model is used to actively select $\vec{x}_i$ to acquire a label for \citep{kirsch2022unifying}.
Once the label is acquired, the sample is added to the training set. 
In the pool, $\mathcal{D}^{\text{pool }}$, the censorship status of new observations is \emph{unknown}, i.e., during acquitions of new observations, both $y_i$ and $\ell_i$ are unknown~\citep{zafar2019adeepactive}. 
Thus, acquiring new labels involves obtaining its label $y_i$ and its censorship status $\ell_i$~\citep{vinzamuri2014active}. 

\subsection{Bayesian Experimental design}
Bayesian experimental design is a formal framework for quantifying the information gained from an experiment~\citep{lindley1956onameasure}. 
In active learning, we can view the input $\vec{x}_i$ as the design of an experiment and the acquired label $y_i$ as the experiment's outcome and formalise the information gained from observing $y_i$~ \citep{brickford2023preds}.
%\textcolor{myblue}{
%In this paper, we also consider
%in the following the minus sign is not used. the reason is as follows. the maximum information in a stastics sense, will be obtained when when the probability distribution is concentrated on a single value, and information will be reduced as the distribution spreads. This is opposite of what we want, therefore we reverse the scale \citep{lindley1956onameasure}
%}

Let $\theta$ be the quantity we are trying to infer.
Given a prior (or the most recent knowledge), $p(\theta)$, and a likelihood function, $p(y_i | \vec{x}_i, \theta)$, then we can quantify the information gain ($\operatorname{IG}$) in $\theta$ due to an acquisition of $(\vec{x}_i,y_i)$, as the reduction in Shannon entropy in $\theta$ that results from observing $(\vec{x}_i,y_i)$:
\begin{equation}
    \operatorname{IG}_\theta(\vec{x}_i, y_i) = \operatorname{H}[p(\theta)] - \operatorname{H}[p(\theta | \vec{x}_i, y_i)]\, .
\end{equation}
Since $y_i$ is a random variable, the expected information of $y_i$, can be computed across multiple simulated outcomes, using
\begin{equation}
    p_\theta(y_i | \vec{x}_i) = \mathbb{E}_{p(\theta)}[p(y_i| \vec{x}_i,\theta)], 
\end{equation}
which leads to the expected information gain,
\begin{equation}
    \operatorname{EIG}_{\theta}(\vec{x}_i) = \mathbb{E}_{p_\theta(y_i| \vec{x}_i)}\left[\operatorname{H}\left[p(\theta)\right] - \operatorname{H}\left[p(\theta | \vec{x}_i, y_i)\right]\right]\,.
\end{equation}
This is the expected reduction in uncertainty of $\theta$ after conditioning on $(\vec{x}_i, y_i)$. Equivalently it is the mutual information between $\theta$, and $y_i$ given $\vec{x}_i$, denoted $\operatorname{I}(y_i, \theta | \vec{x}_i)$~\citep{brickford2023preds}.

\subsection{Bayesian active learning}
The expected information gain has often been the basis for Bayesian active learning, seeking to acquire data points that provide high information gain in the model parameters $\theta$.
This acquisition function is referred to as the \emph{Bayesian Active Learning by Disagreement} (BALD)~\citep{houlsby2011bayesian}:
\begin{equation}
\label{eq:bald}
\begin{aligned}
\text {BALD} &\left(\vec{x}_i\right) =  \operatorname{I}(y_i, \theta | \vec{x}_i) \\
&= \mathbb{E}_{p_\theta(y_i| \vec{x}_i)}[\operatorname{H}[p(\theta)] - \operatorname{H}[p(\theta| \vec{x}_i, y_i)]] \\
&= \mathbb{E}_{p(\theta)}[\operatorname{H}[p_\theta(y_i| \vec{x}_i)] - \operatorname{H}[p_\theta(y_i | \vec{x}_i, \theta)]]\, .\\
\end{aligned}
\end{equation}
The BALD score is often used when the update to the model parameters is non-Bayesian, for example, when applying Monte Carlo dropout in a neural network \citep{gal2017deep}.
For Bayesian active learning without censoring, the $\text{BALD}(\vec{x}_i)$ acquisition function can be used for classification and regression methods, as the entropies are well-defined for these tasks \citep{gal2017deep, jesson2021causal}.

\section{Censoring and Information}
Ideally, we would still like to use the $\text{BALD}$ objectives for the censored regression case. 
However, we must consider that, for a new observation $\vec{x}_i$ in the pool, the corresponding label $y_i$ can provide a varying amount of information for the distribution of $y^*_i$ and $\theta$, depending on the censorship status of the label \citep{baxter1989anote, hollander1985information, hollander1987measuring, hollander1990information}.
The censorship status of new observations, $\vec{x}_i$, is unknown before we acquire the label, which means that the information provided by the label is unknown at the time of acquisition.

To use the $\operatorname{EIG}$ and BALD acquisition functions, we will derive the Shannon entropy for a model trained with Equation \ref{eq:tobit_fit} and adopt Bayesian experimental design ideas to reason about the information gain that $y_i$ provides to the true distributions of $y^*_i$.
Using the derived entropy, we extend the BALD objective to the censored case and use this as an acquisition function for Bayesian active learning in this setting. 
For the entropy equations in the following, we omit the dependency on $\vec{x}_i$ and $\theta$ for readability.

\subsection{Information of two experiments}
We consider the acquisition of a new label $y$, where $y$ can come from two different probability distributions.
Following \citet{lindley1956onameasure}, the information that $y$ holds (i.e., Shannon entropy) is either the entropy from the first or the second probability distributions.

\paragraph{Definition 1.}  \textit{
If $p(y)$ is a mixture of two distributions, that is, $p(y)=\lambda p_1(y) + (1-\lambda)p_2(y)$  with $\lambda$, $0 < \lambda < 1$, then observation $y$ is from to the density $p_1(y)$ with probability $\lambda$ and with probability $1-\lambda$, $y$ is from $p_2(y)$. 
The the entropy of $y$ is }
%\begin{equation}
    %\operatorname{H}[y]= \lambda H\left[p_1(y)\right] + (1-\lambda) H\left[p_2(y)\right]\, .
 %   \operatorname{H}[\lambda p_1(y) + (1-\lambda)p_2(y)]= \lambda H\left[p_1(y)\right] + (1-\lambda) H\left[p_2(y)\right]\, .
% H\left[\lambda p_1(y) + (1-\lambda) p_2(y)\right] 
%=\lambda H\left[p_1(y)\right] + (1-\lambda) H\left[p_2(y)\right]\,
\begin{equation}
%\end{equation}
\begin{aligned}
H[p(y)] &= H [ \lambda p_1(y) + (1-\lambda)p_2(y) ] \\
&= \lambda H\left[p_1(y)\right] + (1-\lambda) H\left[p_2(y)\right].\
%&\lambda H\left[p_1(y)\right] + (1-\lambda) H\left[p_2(y)\right]
%\operatorname{H}\left[p(y)\right]= &H\left[\lambda p_1(y) + (1-\lambda) p_2(y)\right] \\
%=&\lambda H\left[p_1(y)\right] + (1-\lambda) H\left[p_2(y)\right]\,
%\operatorname{H}\left[ y \right] = \operatorname{H}[\lambda \varphi(y) + &(1-\lambda) (1-\Phi(z))] \,
\end{aligned}
\end{equation}
We refer the reader to \citet{lindley1956onameasure} for theoretical analysis of this information measure.

\subsection{Censored information}
In censored regression, the censoring status of new data points, $\vec{x}_i$, from the pool, \dpool, is unknown, and we do not know if $y=y^*$ or $y=z$.
Therefore, an observed label $y$ will have varying entropy about $p(y^*)$ depending on its censoring status.
This case is analogous to Definition~1, where $y$ is either obtained according to the distribution $\varphi$ if $y$ is uncensored, or from the censoring distribution, $(1-\Phi)$, if $y$ is censored.
We consider the entropy of $y^*$ as:

\paragraph{Proposition 1.} \textit{(Information of censored experiments)
Let $\lambda$ be the probability that a new observation is censored, i.e.~$p(\ell=1) = \lambda$.
Then with probability $\lambda$, $y$ is an uncensored value and is from the density $\varphi(y)$. With probability ($1-\lambda)$, the observation is censored, in which case the probability density of $y$ is $(1-\Phi(z))$.
%If we are informed about $y$ and if the observation is censored, then 
The entropy of $y$ when it is subject to censoring is,}
\begin{equation}
    %\operatorname{H}[y]= \lambda H\left[\varphi(y)\right] + (1-\lambda) H\left[(1-\Phi(z))\right]\, .
    \operatorname{H}[p(y)]= H\left[\lambda \varphi(y) + (1-\lambda) (1-\Phi(z))\right]\, .
\end{equation}

\paragraph{Proof}
In the case of non-censorship, the entropy of $y$ and the continuous distribution $p(y)$ corresponds to the Shannon differential entropy, defined as,
\begin{equation}
    \operatorname{H}[p(y)]= - \int p(y)\log p(y) dy\, .
\end{equation}
%However, given the right censorship at z, y is defined on the measurable space
However, given the right censorship at $z$, $y$ is defined on the measurable space $(-\infty; z]$, and we split the integral at the censoring threshold $z$ into an uncensored and censored case.
For the case where the observation $y$ is not censored and in the set $(-\infty; z]$, then $p(y)=\varphi(y)$ and if $y^* > z$, then the observations are censored and $p(y)=1-\Phi(z)$.
Therefore, we can rewrite the entropy as,
\begin{equation}
\begin{aligned}
    \operatorname{H}[p(y)]= &- \int_{-\infty}^\tau p(y^*) \log\varphi(y^*) \,dy^*\\ &- \int_\tau^\infty  p(y^*) \log(1-\Phi(z)) \,dy^*\, .
    \end{aligned}
\end{equation}

If we assume that $z$ is known (or constant), we can approximate this integral with Monte Carlo samples of $y^*$ as,
\begin{equation}
\begin{aligned}
\operatorname{H}[p(y)]=-\mathbb{E}_{y^* \sim p\left(y^*\right)}[&\mathbbm{1}\{y^*\leq z\}] \log \varphi\left(y^*\right)+ \\
& \mathbbm{1}\{y^* > z\} \log (1-\Phi(z))]\, .
\end{aligned}
\end{equation}
The entropy can be interpreted as if we know the censoring; we know how much information we can expect from a new observation.
However, the censoring threshold is \emph{unknown}, and instead of a hard assignment, we can consider the probability $\lambda$ of an observation being censored or not.
This leads to the following entropy, 
\begin{equation}
\label{eq:al_entropy}
\begin{aligned}
\operatorname{H}\left[p(y)\right]=-\mathbb{E}_{y^* \sim p\left(y^*\right)}[&\lambda \log \varphi\left(y\right)+ \\
& (1-\lambda) \log (1-\Phi(z))]\, .%\\
%\operatorname{H}\left[ y \right] = \operatorname{H}[\lambda \varphi(y) + &(1-\lambda) (1-\Phi(z))] \,
\end{aligned}
\end{equation}
The entropy can be interpreted as follows: if we know the censoring status, we understand how much information to expect from a new observation~\citep{hollander1990information}.

\subsection{Conditional entropy}
The entropy derived in Equation \ref{eq:al_entropy}, involves the expectation with respect to the distribution of $p(y^*)$, and is essentially a form of conditional entropy by conditioning on the censorship status of $y$.
The entropy captures the uncertainty in the mixture distribution, considering the influence of the censorship status.
It is, therefore, easy to show that this leads to the following entropy, 
\begin{equation}
\label{eq:entropy}
\begin{aligned}
\operatorname{H}\left[p(y|\ell)\right]=-\mathbb{E}_{y^* \sim p\left(y^*\right)}[&p(\ell) \log \varphi\left(y\right)+ \\
& (1-p(\ell)) \log (1-\Phi(z))]\, .%\\
%\operatorname{H}\left[ y \right] = \operatorname{H}[\lambda \varphi(y) + &(1-\lambda) (1-\Phi(z))] \,
\end{aligned}
\end{equation}
We refer the reader to~\citet{hollander1985information, hollander1987measuring, hollander1990information} for a theoretical analysis and a discussion of the entropy with censored data.

\section{Expected information gain in censored acquisitions}
We can use the derived entropy to calculate the information that new observed targets $y_i$ provide to the distribution of $y^*$.
However, the acquisition of new labels $y_i$ not only requires obtaining new values of $y_i$, but it also involves acquiring new censoring indicators $\ell_i$ \citep{zafar2019adeepactive}. Consequently, it is necessary to account for the mutual information between $y_i$ and $\theta$ and consider the information provided by $\ell_i$. As a result, we jointly compute the mutual information between $(y_i,\ell_i)$ and $\theta$.
This leads to the following mutual information, 
\begin{equation}
\label{eq:cbald}
    \begin{aligned}
        \mathcal{C}&\text{-BALD}(\vec{x}_i) =\operatorname{I}\left[(y_i, \ell_i), \theta | \vec{x}_i\right] \\
        &= \mathbb{E}_{p(\theta)}[\operatorname{H}[p_\theta(y_i, \ell_i| \vec{x}_i)] - \operatorname{H}[p(y_i, \ell_i | \vec{x}_i, \theta)]] \, .\\
 %   \end{aligned}
%\end{equation}
%This mutual information simplifies to,
%\begin{equation}
%\label{eq:cbald}
%\begin{aligned}
    %\mathcal{C}\text{-BALD}(\vec{x}_i) 
    %&= \operatorname{I}\left[y_i,\ell_i, \theta | \vec{x}_i\right] \\
    & = \operatorname{I}\left[y_i, \theta |\ell_i, \vec{x}_i\right] +\operatorname{I}\left[\ell_i, \theta | \vec{x}_i\right] \, .\\
\end{aligned}
\end{equation}

%Under the assumption that the censoring is independent given the covariates (Assumption 1) \citep{Subramanian2006hh}, the mutual information is simplified to,
%\begin{equation}
%\label{eq:cbald}
%\begin{aligned}
%    \mathcal{C}\text{-BALD}(\vec{x}_i) &= \operatorname{I}\left[y_i,\ell_i, \theta | \vec{x}_i\right] \\
%    & = \operatorname{I}\left[y_i, \theta | \vec{x}_i\right] +\operatorname{I}\left[\ell_i, \theta | \vec{x}_i\right] \, .\\
%\end{aligned}
%\end{equation}
We provide the proof in the Appendix \ref{sec:proof_cbald}.

In the censored regression, the information gained from observing $y_i$ and $\ell_i$ is the information provided by observing the label $y_i$ given the censoring indicator $\ell_i$, plus the information from observing the censoring indicator $\ell_i$.
The mutual information criteria can be computed as the BALD objectives,
%\begin{equation}
\begin{align}
    \operatorname{I}\left[y_i, \theta | \ell_i, \vec{x}_i\right] &= \mathbb{E}_{p(\theta)}[\operatorname{H}[p_\theta(y_i|\ell_i, \vec{x}_i)] - \operatorname{H}[p(y_i | \ell_i, \vec{x}_i, \theta)]]\, , \\
    \operatorname{I}\left[\ell_i, \theta | \vec{x}_i\right] & = \mathbb{E}_{p(\theta)}[\operatorname{H}[p_\theta(\ell_i| \vec{x}_i)] - \operatorname{H}[p(\ell_i | \vec{x}_i, \theta)]] \, .
\end{align}
%\end{equation}a

\subsection{Modelling approach}
\begin{figure}[tb]
\centering
%\resizebox{0.49\textwidth}{!}{%
\begin{tikzpicture}
  %\tikz{
% nodes
%\clip(-4.3,-10.66) rectangle (17.58,6.3);
 \node[obs] (x) {$\vec{x}_i$};%
 \node[latent,right=of x, fill] (yt) {$y_i^*$}; %
 \node[latent,right=of x, yshift=-1cm] (z) {$z_i$}; %
 \node[obs, right=of yt] (y) {$y_i$};%
 %\node[latent,above=of yt, xshift=-0.5cm, fill] (ms) {$\mu^*$}; %
 %\node[latent,above=of yt, xshift=0.5cm,yshift=-0.5cm, fill] (ss) {$\sigma^*$}; %
 %\node[latent,right=of y, yshift=-0.5cm, fill] (m) {$\mu$}; %
 %\node[latent,right=of y, yshift=0.5cm, fill] (s) {$\sigma$}; %
 \node[obs,below=of y] (l) {$\ell_i$}; %
 %\node[factor,right=of y,xshift=1cm, label=$\theta$] (tt) {}; %
 \node[latent,right=of y] (tt) {$\theta$}; %
 %\factor[latent,right=of y] (tt) {$\theta$}; %
 %\node[latent,right=of l, fill] (lam) {$\lambda$}; %
 \plate [inner sep=.25cm,yshift=.2cm] {plate1} {(x)(y)(z)(l)} {$N$}; %
% edges
 %\edge {x} {yt,z}  
 \path (x) edge[->] (yt);
 \path (x) edge[->] (z);
 %\edge {yt,z, l} {y} 
 \path (yt) edge[->] (y);
 \path (l) edge[->] (y);
 \path (z) edge[->] (y);
 %\edge {ms, ss} {yt}  
 %\edge {m, s} {y}  
 %\edge {lam} {l}  
 \path (x) edge[bend right,dashed, ->] (l) ;
 \path (x) edge[bend left,dashed, ->] (y) ;
 \path (yt) edge[right, ->] (l) ;
 \path (z) edge[left,->] (l) ;
 %\path (tt) edge[bend right,dashed, ->] (ms) ;
 %\path (tt) edge[bend right,dashed, ->] (ss) ;
 \path (tt) edge[bend right,dashed, myred, ->] (yt) ;
 \path (tt) edge[->,dashed, myblue] (y) ;
 \path (tt) edge[->, dashed,myorange] (l) ;
 %\factoredge{tt}{l}
 %}
 \end{tikzpicture}
 %}
    %\caption{model fit }
    %\label{fig:model_fit}
\caption{Overview of the modelling approach. We propose to model the distributions \textcolor{myorange}{$p_\theta(\ell_i| \vec{x}_i)$}, \textcolor{myblue}{$p_\theta(y_i | \vec{x}_i)$}, and \textcolor{myred}{$p_\theta(y^*_i| \vec{x}_i)$}. Shaded circles indicate observed values and unshaded circles indicate latent values. 
%$\theta$ is stochastic sampled from the posterior $p(\theta | \mathcal{D})$.
}
\label{fig:pgm}
\end{figure}
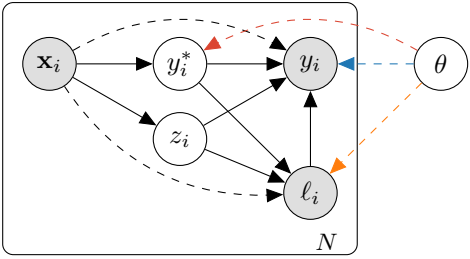

A fundamental limitation of \scbald for active learning is that the primary model from Equation~\ref{eq:tobit_fit} only approximates the distribution $\varphi$ and $\Phi$.
This means that during acquisition, there is no knowledge of the potential censoring status of new observations $\ell_i$, which should be used in the mutual information (Equation \ref{eq:cbald}), and there is no knowledge of the potential censoring threshold $z_i$, which should be used to compute the entropy of $y_i$ (Equation \ref{eq:entropy}).
Therefore, applying \scbald in practice is not straightforward.
To overcome these challenges, we propose explicitly modelling the probability of being censored $\ell_i$ and the censoring threshold $z_i$, which we describe below.
Figure \ref{fig:pgm} provides an overview of the proposed modelling approach.

\paragraph{Modelling of $\mathbf{\ell_i}$:} 
Recall that the censoring indicator $\ell_i = \mathbbm{1}\{y^*_i\leq z_i\}$ is observed for each data point in a censored dataset.
It is a binary indicator of whether the observations are censored or not.
We propose to approximate the distribution of $p(\ell_i | \vec{x}_i, \theta)$. 
We parameterise $p(\ell_i | \vec{x}_i, \theta)$, as a Bernoulli distribution, and infer the parameters $\theta$ using the binary cross entropy ($\mathcal{L}_{\text{BCE}}(\theta)$). 
We approximate the probability of censoring new observations using this distribution, i.e., $\lambda_i$.
Consequently, this explicit modelling of $\ell_i$ allows us to compute the mutual information $\operatorname{I}[\ell_i, \theta | \vec{x}_i]$.

\paragraph{Modelling of $\mathbf{z_i}$:} 
Explicit modelling of $z_i$ is more challenging, as it is not fully observed (similar to  $y^*_i$)
Recall that we observe, $y_i=\min(y^*_i,z_i)$, i.e. when an observation is censored, we observe $z_i$ or if it is not censored, $y^*$.
We only observe $z_i$ when an observation is censored. 
To approximate $z_i$, we propose approximating the continuous distribution of $p(y_i | \vec{x}_i, \theta_i)$ for new observations.
We use $p(y_i | \vec{x}_i, \theta_i)$ to estimate the \emph{observed} values, $\hat{y}_i$, for new data points in the \dpool.
%Recall that the observed $y_i$ equals $y^*_i$ when an observation is not censored and equals $z$ when censored. 
%We propose approximating the continuous distribution of $p_\theta(y | \vec{x}_i)$.
%The expected value from this distribution can be used as a proxy for $z_i$ when the observations are censored.

\paragraph{Entropy estimation:}
With this explicit modelling approach, we can approximate the entropy from the observation $y$ to the distribution $y^*$,
\begin{equation}
\label{eq:final_entr}
\begin{aligned}
\operatorname{H}[p(y_i|\ell_i)]=& -\mathbb{E}_{y^*_i \sim p\left(y_i\right)}[\textcolor{myorange}{p_\theta(\ell_i | \vec{x}_i)} \log \textcolor{myred}{\varphi}\left(\textcolor{myblue}{\hat{y}} \right)  +\\
&(1-\textcolor{myorange}{p_\theta(\ell_i | \vec{x}_i)}) \log (1-\textcolor{myred}{\Phi}(\textcolor{myblue}{\hat{y}}))]\, .
\end{aligned}
\end{equation}

\subsection{Summary and implementation details}
We want to use the mutual information between observations of $y_i$, $\ell_i$, and the model parameters $\theta$ to acquire new labels to reduce model uncertainty about $y^*_i$.
Since the distribution of $y^*$ is not fully observed, we use the entropy defined in Equation \ref{eq:entropy} to compute the mutual information.
%In practice, we also reverse the scale of the entropies. The reason is as follows: the maximum information , in a statistical sense, will be obtained when the probability distribution is concentrated on a single value, and information will be reduced as the distribution spreads. This is the opposite of what we want; therefore, we reverse the scale \citep{lindley1956onameasure}.
However, the entropy relies on the knowledge of unknown variables $z_i$ and $\ell_i$. We propose to model them explicitly, resulting in the estimated entropy of Equation \ref{eq:final_entr}.
\begin{figure}[tb]
    \centering
    \includegraphics[width=\columnwidth]{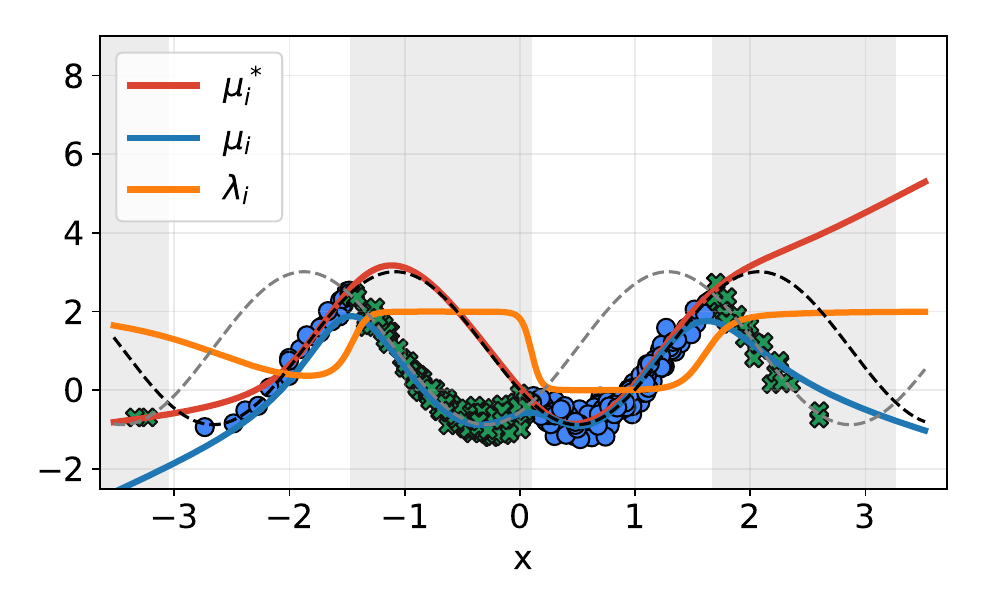}
    \caption{Overview of the fit of the proposed modelling approach on the 1-D synthetic dataset. Grey areas indicate where $y^* \leq z$, and the black line is the true function we are trying to approximate.
    \textcolor{myred}{Red}: Estimated distribution of the true function.
    \textcolor{myblue}{Blue}: Estimated distribution of the observed values.
    \textcolor{myorange}{Orange}: Estimated probability of being censored (scaled between 0 and 2 for illustration purposes.)
    }
    \label{fig:model_fit}
\end{figure}

\begin{figure*}
%\centering
%\includegraphics[width=0.9\columnwidth]{AISTATS2024PaperPack/figures/fit_cbald_0_33.pdf}
%\includegraphics[width=0.9\columnwidth]{AISTATS2024PaperPack/figures/scores_cbald_0_35.pdf}
    %\includegraphics[width=0.33\textwidth]{figures/fit_cbald_0_27.pdf   }
    %\includegraphics[width=0.33\textwidth]{figures/scores_cbald_0_27.pdf}
    %\includegraphics[width=0.33\textwidth]{figures/synth_performance.pdf}
    \includegraphics[width=0.50\textwidth]{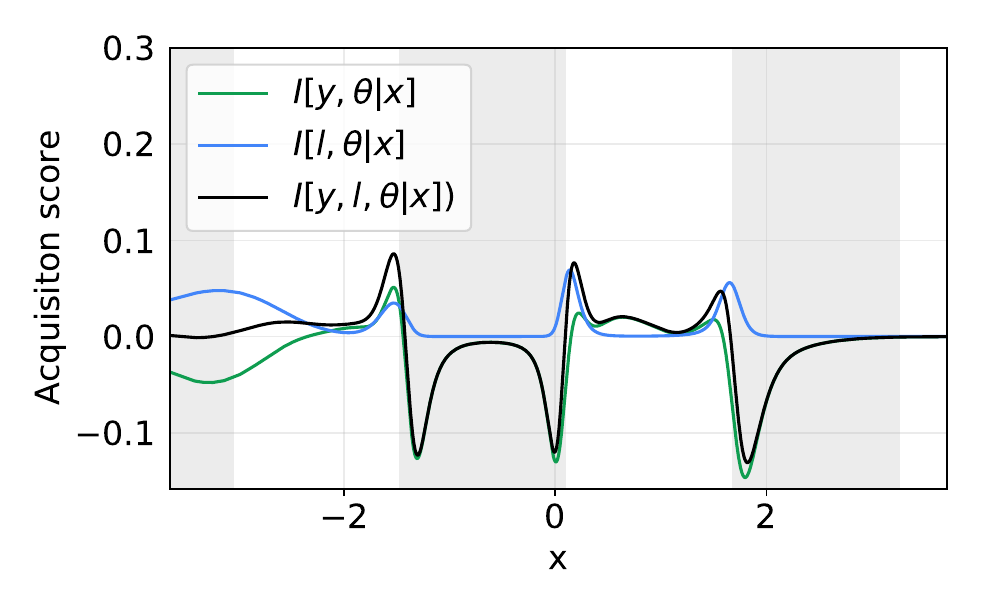}
    \includegraphics[width=0.50\textwidth]{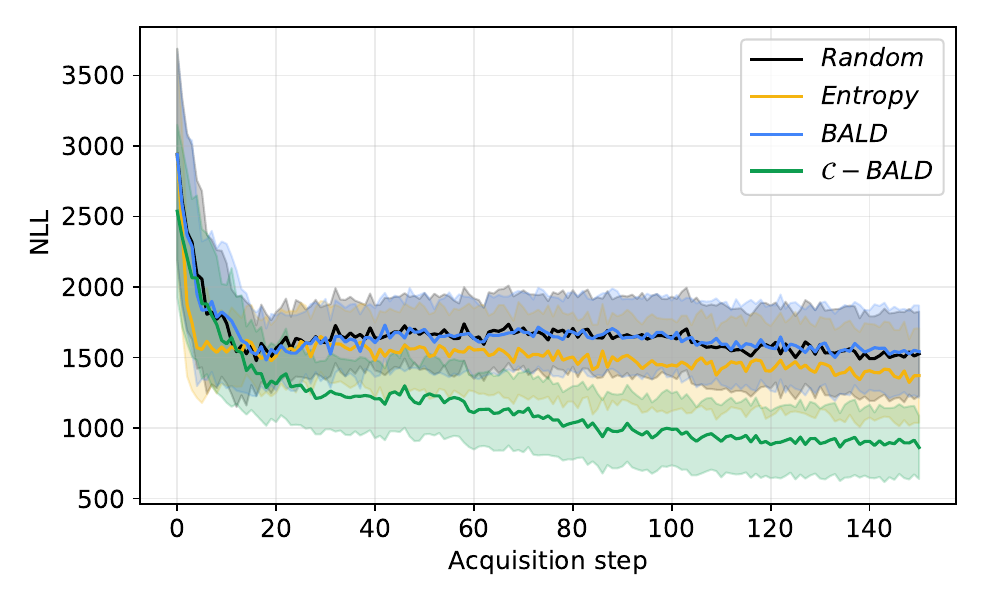}
    
    \caption{\textbf{Left):} The mutual information calculations for the label $y$ and the censoring status. 
    Grey areas indicate areas with complete censoring.
    Most information comes from the cross-over point between the censored and non-censored values.
    \textbf{Right):} The right censored NLL for the models across different acquisition functions on the synthetic dataset (mean $\pm$ standard error). \scbald achieves the best overall fit on the test set.}
    \label{fig:synthetic_data_experiment}
\end{figure*}

\textbf{Implementation:}
We will use Gaussian distributions for $y^*_i$ and $y_i$ and a Bernoulli distribution for $\ell_i$.
When $y^*_i$ is assumed to be Gaussian, the model corresponds to a Tobit model \citep{tobin1958estimation}.
We enforce the constraint that $\sigma^*_i$ and $\sigma_i$ should be positive by applying the softplus activation function on these parameters.
The parameters of the Gaussian $p(y_i | \vec{x}_i, \theta)$ are approximated with the maximum likelihood of the parameters ($\mathcal{L}_{\text{MLE}}(\theta))$.
To summarise,
%\begin{equation}
\begin{align}
    p(y^*_i| \vec{x}_i, \theta) &\sim \underbrace{\mathcal{N}(\mu_i^*, \sigma_i^{2*} | \vec{x}_i,  \theta)}_{{\textcolor{myred}{\text{True distribution of $y^*_i$.}}}}\, , \\
    p(y_i | \vec{x}_i, \theta) &\sim \underbrace{\mathcal{N}(\mu_i, \sigma_i^2| \vec{x}_i, \theta)}_{{\textcolor{myblue}{\text{Distribution of $y_i$.}}}}\, , \\
    p(\ell_i|\vec{x}_i, \theta) &\sim \underbrace{\operatorname{Ber}(\lambda_i | \vec{x}_i, \theta)}_{{\textcolor{myorange}{\text{Distribution of $\ell_i$.}}}}\, . 
\end{align}
%\end{equation}
We model all these distributions with a single Bayesian neural network with stochastic parameters. The output of the Bayesian Neural network is the distributional parameters of the distributions $p(y_i^*| \vec{x}_i, \theta)$, $p(y_i| \vec{x}_i, \theta)$ and $p(\ell_i | \vec{x}_i, \theta)$, i.e. five outputs neurons for the set $\{\mu_i^*,\sigma_i^*,\mu_i,\sigma_i,\lambda_i\}$.

The parameters of the neural network, $\theta$, are inferred using the total loss from the maximum likelihood estimation of all these distributions, 
\begin{equation}
    \mathcal{L}(\theta) = \mathcal{L}_{\mathcal{C}}(\theta) + \mathcal{L}_{\text{MLE}}(\theta) + \mathcal{L}_{\text{BCE}}(\theta)\, .
\end{equation}

Figure \ref{fig:model_fit} shows the fit of the proposed model for all the different distributions on a synthetic dataset.
Using all the explicit modelling of $y^*_i$, $y_i$ and $\ell_i$, we can compute the \scbald objective and use it as an acquisition function in active learning.

\section{Experiments}
In this section, we present the results of the proposed acquisition function with multiple experiments on synthetic and real-world datasets.
%The main objective is to show that the proposed acquisition function identifies which points provide the most information, leading to better model fits and superior acquisition of new data points. 

\textbf{Models:}
We implement the Bayesian Neural Network with stochastic parameters using Monte Carlo Dropout \citep{gal16dropout}.
We use three layers, 128 hidden units, a dropout probability of 0.25, and the ADAM optimiser with a learning rate of $0.3\cdot10^{-3}$~\citep{kingma2014adam} and the ReLU activation function\footnote{In Appendix \ref{sec:size_experiments}, we experiment with different model architectures.}.

\textbf{Baselines:} We compare the proposed acquisition function with the following baselines: \textbf{Random} acquisitions, which randomly acquires data points in \dpool, the \textbf{Entropy} (\sunc) of  Bayesian neural networks, which is proportional to variance between the individual's models in the sampled ensemble, $\operatorname{Var}_{\theta \sim p(\theta|\mathcal{D})}[p(y_i| \mathbf{x}_i)]$, and the \textbf{BALD} objective from Equation \ref{eq:bald}.

\begin{table*}[tb]
%\vskip 0.15in
\begin{center}
\begin{small}
\begin{sc}
\resizebox{\textwidth}{!}{
\begin{tabular}{lrr|rrrrrrr}
\toprule
Name & Num. features & Censorship & $n_0$ & Acquisition size& Acquisition steps & repetitions & \dpool & \dval & \dtest \\
\midrule
Synthetic & 1 & $44\%$ & 10 & 3 & 150 & 50  & 9000 & 250 & 500 \\
BreastMSK & 5& $77\%$& 5 & 3 & 150& 50 & 1285 & 183 & 366 \\
Metabric & 9& $42\%$ & 5 & 3 & 150& 50 & 1523 & 76 & 305 \\
Whas & 6 & $58\%$ & 5 & 3 & 150& 50 & 1310 & 65 & 263 \\
GBSG & 7& $37\%$& 5 & 3 & 150& 50 & 1546 & 137 & 549 \\
Support & 14& $32\%$ & 5 & 3 & 150& 50 & 7098 & 355 & 1420 \\
Churn & 26 & $ 53\%$ & 5 & 3 & 150& 50 & 1276 & 136&  546\\
Credit Risk & 47 & $30\%$  & 5 & 3 & 150& 50 & 650 & 70 &  280\\
SurvMNIST  & $28\times28$ & $53\%$ & 100 & 5 & 100 & 25 & 60000 & 5000 & 5000 \\
\bottomrule
\end{tabular}
}
\end{sc}
\end{small}
\end{center}
\caption{Overview of the various datasets used in this analysis, including the number of features and the percentage of censorship in \dpool. We also include $n_0$ as the initial data points in \dtrain}
\label{tab:dataset_overview}
%\vskip -0.25in
\end{table*}

\textbf{Evaluation:} 
We evaluate the acquisition function based on the method outlined by \citet{riis2022bayesian}.
To quantify the performance of the acquisition function, we evaluate the relative decrease in the area under the curve (RD-AUC) across the entire active learning experiment. 
We compare the relative decrease to a baseline acquisition function (Random) and evaluate the models' right censored negative log-likelihood (NLL) on a test set (\dtest).
Since the NLL is not bounded by 0, we use the lowest NLL obtained across all the acquisition functions as a lower bound for the metric.
We compute the average across all the number of acquisitions, $N_\text{Acq}$.
The RD-AUC is defined as follows:
\begin{equation}
\text{RD-AUC} = \frac{1}{N_\text{Acq}}\sum_{i=0}^{N_\text{Acq}}\left(\frac{NLL_{\text{Random}}-NLL_{s}}{NLL_{\text{Random}}}\right)\, ,
\end{equation}
where $NLL_s$ is the negative log-likelihood of the model with the acquisition function $s$ and $NLL_{\text{Random}}$ is the negative log-likelihood of from Random acquisition. 

\textbf{Synthetic Data:}
We begin our empirical evaluation of the proposed acquisition by considering the following 1D synthetic dataset, with $x_i = \mathcal{N}(5,1)$, and,
\begin{align}
y_i^* &= \frac{1}{2}\sin(2x_i) + 2 + \varepsilon_i\, ,\\
z_i &= \frac{1}{2}\cos(2x_i) + 2 + \varepsilon_i\, ,
%y_i &= \min(y^*_i, z_i)\, ,
\end{align}
$y_i = \min(y^*_i, z_i)$, and $\ell_i = \mathbbm{1}\{y^*_i \leq z_i\}$ and $\varepsilon_i\sim\mathcal{N}(0, 0.01 |x_i|)$. The dataset is depicted in Figure \ref{fig:dataexample} and our proposed modelling fit in Figure \ref{fig:model_fit}.
We generate a small pool of labelled data points ($n_0=10$), a larger set of unlabelled data points $|$\dpool$|=9000$, and a $|$\dtest$|=500$.

We train a model of the small pool of labelled data and acquire three new data points with labels every iteration. During each training step, we use a small validation set \dval with 250 observations to evaluate the models and apply early stopping on the right censored maximum likelihood. 

Figure \ref{fig:synthetic_data_experiment} shows the \scbald scores across the entire range of $x$. 
$\mathcal{C}$-BALD assigns a high mutual information value in regions where the censoring status changes, i.e. when the model is uncertain about the information that new samples will provide. 
In the right of Figure \ref{fig:synthetic_data_experiment}, we show the right censored negative log-likelihood for the different acquisition functions.
We find that \scbald achieves the best overall fit of the data with the lowest NLL, which shows that it identifies which data point provides the most information to the model. 

\textbf{Real Datasets:}
We test the proposed functions on seven real-world datasets: five from a biomedical context~\citep{Katzman2018deepsurv} and two from a predictive analytics context~\citep{pysurvival_cite}.
Three datasets focus on estimating the survival time for various types of cancer patients (\textbf{BreastMSK}, \textbf{METABRIC}, and \textbf{GBSG}), one dataset for modelling the survival time of myocardial infarction (\textbf{WHAS}), and the last dataset estimates the survival time for critically-ill hospital patients (\textbf{SUPPORT}).
For the predictive analytics datasets, we focus on predicting the time customers remain subscribed to a service (\textbf{Churn}) and the other on estimating the time for borrowers to repay their credit (\textbf{Credit Risk}).
These datasets contain \emph{real} censoring, i.e., no synthetic censoring is applied to them\footnote{A more extensive summary of these datasets can be found in Appendix \ref{sec:real_datasets}.}.

Table~\ref{tab:dataset_overview} summarises the datasets used in the experiments, including the number of features, the percentage of censored observations, and the total number of observations. Additionally, it includes a summary of the parameters used for the active learning experiments for each dataset.
The results reported are averages over the number of repetitions for each dataset and acquisition function (mean $\pm$ standard error).
%While the test set includes censored data points, the NLL from Equation \ref{eq:tobit_fit} is a valid scoring function~\citep{rindt2022survival}.

\begin{figure*}[tb]
    %\centering
    \includegraphics[width=0.49\textwidth]{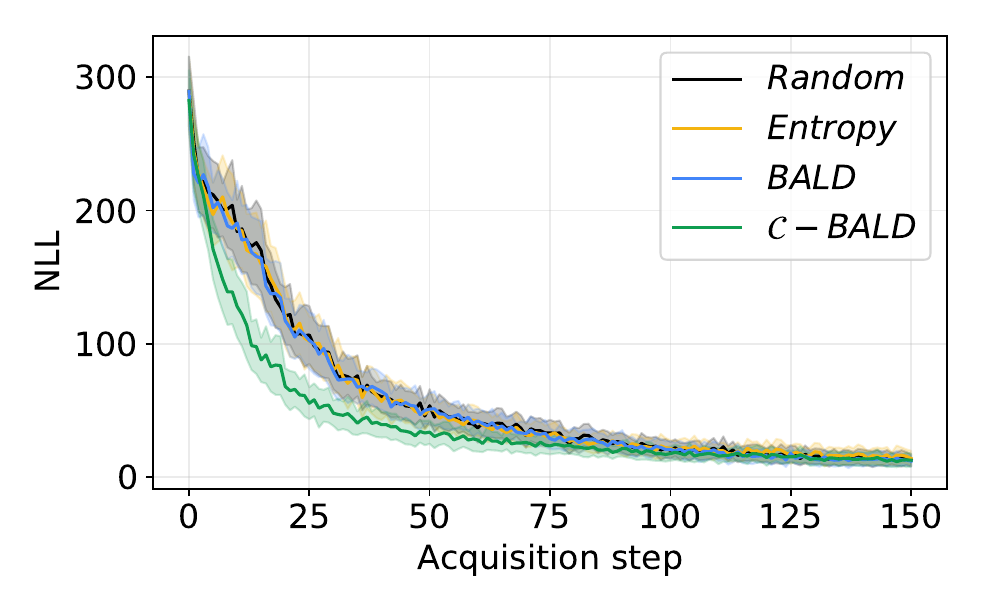}
    \includegraphics[width=0.50\textwidth]{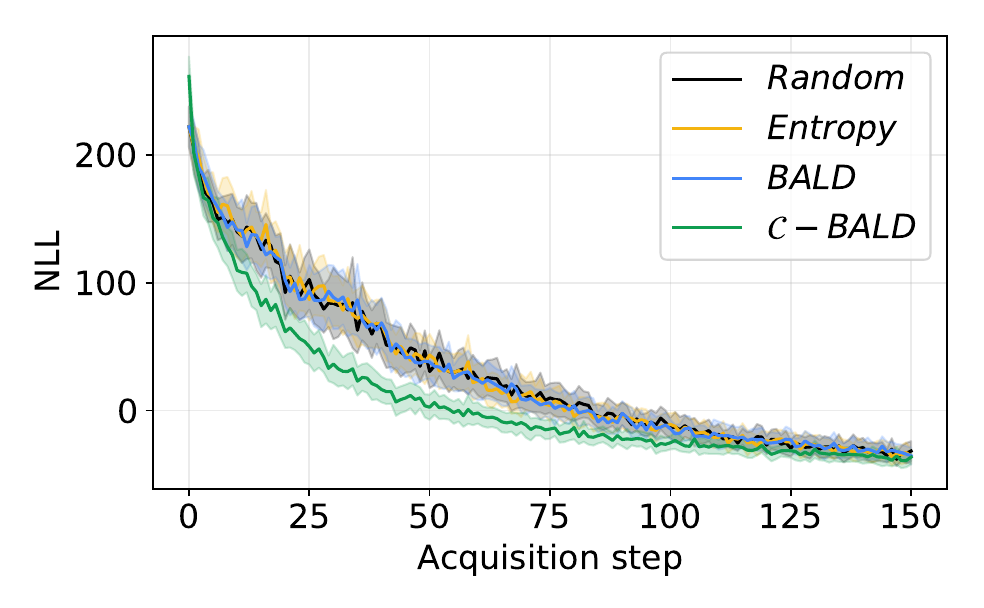}
    \caption{Results of the real-world experiments on two of the seven datasets, namely the METABRIC and CREDIT RISK datasets, respectively. The figure shows the NLL (mean $\pm$ standard error) across the multiple repetitions of the experiment.}
    \label{fig:real_results}
\end{figure*}

Table \ref{tab:relative_dcrease_AUC} reports the RD-AUC compared across the different scoring functions.
Figure \ref{fig:real_results} shows the right-censored NLL across the different runs for two real-world datasets.
%We find that the proposed acquisition function obtains a superior fit compared to the other acquisition strategies.
We find that the proposed acquisition function leads to better acquisition of new data points by obtaining a superior fit on the test set compared to the baselines.

\begin{table}[tb]
%\vskip 0.15in
\begin{center}
\begin{small}
\begin{sc}
\resizebox{\columnwidth}{!}{
\begin{tabular}{l|rrrr}
\toprule
Dataset &  \srandom & \sunc &  \sbald & \scbald  \\
\midrule
Synthetic & $0.00 \pm 0.00$ & $8.65 \pm 0.42$ &  $-0.12 \pm 0.14$ & $\mathbf{33.49 \pm  1.11}$ \\
BreastMSK & $0.00 \pm 0.00$ & $8.21 \pm 1.43$ &  $-1.89 \pm 0.66$ & $\mathbf{8.75 \pm  1.42}$ \\
Metabric & $0.00 \pm 0.00$ & $-0.67 \pm 0.39$ &  $2.25 \pm 0.34$ & $\mathbf{18.26 \pm  0.94}$ \\ 
whas &  $0.00 \pm 0.00$ & $0.42 \pm 0.27$ &  $\mathbf{1.68 \pm 0.17}$ & $0.26 \pm  0.32$ \\
GBSG &  $0.00 \pm 0.00$ & $-0.81 \pm 0.05$ &  $-0.04 \pm 0.05$ & $\mathbf{5.58 \pm  0.05}$ \\
support & $0.00 \pm 0.00$ & $0.70 \pm 0.02$ &  $-0.53 \pm 0.01$ & $\mathbf{4.55 \pm  0.02}$ \\
churn & $0.00 \pm 0.00$ & $5.14 \pm 0.31$ &  $0.17 \pm 0.21$ & $\mathbf{32.75 \pm  0.87}$ \\
credit risk & $0.00 \pm 0.00$ & $-0.72 \pm 0.36$ &  $-0.17 \pm 0.33$ & $\mathbf{22.11 \pm  0.64}$ \\
Survmnist & $0.00 \pm 0.00$ & $-0.05 \pm 0.28$ &  $1.06 \pm 0.30$ & $\mathbf{13.47 \pm  0.66}$  \\
\bottomrule
\end{tabular}
}
\end{sc}
\end{small}
\end{center}
\caption{Relative decrease in the area under the curve (RD-AUC) compared to the \srandom scoring function. 
%The values in the table are calculated as the percentage difference between the AUC of the respective scoring function's active learning curve and the active learning curve of Random. 
A higher value in the table represents better performance, with the best performance highlighted in \textbf{bold}.} 
\label{tab:relative_dcrease_AUC}
\vskip -0.15in
\end{table}

\textbf{High-dimensional data:}
Lastly, we evaluate the performance of our proposed scoring functions with Bayesian convolutional neural networks on the \textbf{SurvMNIST} dataset \citep{goldstein2022xcal}. In SurvMNIST, each label is replaced with a random draw from a Gamma distribution, with different distributional parameters across the labels \citep{pearce2022censored}.
%The SurvMNIST dataset was further modified by \citet{pearce2022censored}, where the variance of the gamma distributions varies across different labels. 
The observations in the dataset are censored uniformly, between the minimum and the 90th percentile in the training set \citep{goldstein2022xcal}.
The initial training set contains ten samples from each class in the dataset\footnote{The details of the gamma distributions and the model architecture can be found in Appendix \ref{sec:survmnist_archi}.}.

The experiment on the SurvMNIST dataset shows that the proposed scoring functions outperformed the baseline functions, as shown in Table~\ref{tab:relative_dcrease_AUC}. 

\section{Related work}
The study of the information that an experiment or observation provides was introduced by \citet{lindley1956onameasure} and has often been the basis for new acquisition functions in active learning \citep{mackay1992information, mackay1992theevidence}. 
The study of information in censored experiments has traditionally focused on survival experiments, where observations are studied over time~\citep{hollander1985information, hollander1990information}.
In survival experiments, an individual is observed for an amount of time and is considered censored if the person drops out of the experiment \citep{hollander1987measuring}.
For the discrete and continuous case, the entropy calculations come down to the integral over the time an individual was observed \citep{baxter1989anote}.
\citet{hollander1987measuring} shows that information decreases after censorship and that uncensored observations provide the most information
Traditionally, survival experiments were the focus because conducting such studies in the past was expensive. 
Due to electronic health records, survival analysis today often employs large-scale datasets~\citep{qi2023effective}.
However, in other areas where censored regression is applied, such as transportation systems~\citep{huttel2022modelling, huttel2023mind},
subscription-based businesses \citep{FADER200776, chandar2022usingsurv, maystre2022temporallyconsistent}, and in health survival applications \citep{zafar2019adeepactive, lain2022imageing}, data can be expensive to collect and label, necessitating the need for active learning in this context.

Despite the challenges of censored data, there is limited research on active learning in this context. Two notable exceptions from the survival analysis literature include the work of \citet{vinzamuri2014active}, who proposed a query strategy based on discriminative gradients to identify the most informative points, and the work of \citet{zafar2019adeepactive}, who suggested a query strategy for acquiring data points with the highest expected performance increase if their labels were known.  
%Both of the aforementioned focus on active learning in the parameter space.

A popular approach is Bayesian Active Learning with the BALD objective \citep{houlsby2011bayesian}, specifically with its ability to work in conjunction with deep neural networks~\citep{gal2017deep} and extensions to batch-acquisitions~\citep{kirsch2019batchbald}.
In the Deep Bayesian active learning, the BALD objective has primarily been used for classification tasks with MC Dropout models but has recently seen applications for deep regression tasks, such as estimating causal treatment effects \citep{jesson2021causal} and for black-box models \citep{kirsch2023blackbox}.
While plenty of research has focused on the BALD objective, to our knowledge, we are the first to explore the BALD objective in censored regression.

Our work contributes to Bayesian and Censored regression by extending the common BALD objective to the censored regression case.
We propose a novel modelling approach to approximate the information gain for new observations when they are subject to censoring.
Compared to previous work in active learning with censored regression models, we quantify the information theoretical quantities in the prediction space instead of approximating them in the parameter space.
Future work could explore the implications of censoring on acquisition functions in the parameter space, such as other state-of-the-art methods~\citep{Ash2020Deep, ash2021gone, osti_10336075}, as well as explore other censoring schemes, such as interval-censored data. 
%This paper also focuses on introducing the BALD objective for censored regression. Plenty of research papers have studies this objective, however, the work is focused on 
%and future work could explore how the proposed methods extend to batches \citep{kirsch2019batchbald} or 

\section{Conclusion}
This paper studies Bayesian active learning for censored regression problems.
We have shown that the BALD objective is inadequate for censored data, as the expected information gain for new data points depends on the points' censorship status.
Motivated by this challenge, we derive the entropy for censored distributions and 
propose the $\mathcal{C}$-BALD acquisition function, which accounts for censored observations.
%D and mutual information between new observations and the model's parameters when the observations are subject to censoring.
Since $\mathcal{C}$-BALD relies on unknown variables, we propose to model the variables with minimal computational overhead to compute the mutual information.
Empirically, across synthetic and real datasets, we show that \scbald outperforms BALD with both synthetic and real censoring.
%This suggests that \scbald is a compelling replacement for the BALD objective for datasets subject to censoring.
%EPIG can serve as a compelling drop-in replacement for
%BALD, with particular scope for performance gains when
%using large, diverse pools of unlabelled data.

\begin{acknowledgements}
The research leading to these results has received funding from the Independent Research Fund Denmark (Danmarks Frie Forskningsfond) under the grant no. 0217-00065B.
\end{acknowledgements}

\bibliography{ref}

\onecolumn

\onecolumn

\title{Bayesian Active Learning for Censored Regression\\(Supplementary Material)}
\maketitle

\appendix
%\maketitle
\section{Derivations of the entropy of two independent probability densities.}
Here, we show the joint entropy between two independent distributions $x$ and $y$.
Under the assumption of independence between the distributions, the joint entropy equals the sum of the entropy of $x$ and $y$.
2nd line is under the assumption of independence between $x$ and $y$, and 4th line is the linearity of expectations.
\begin{align}
\operatorname{H}\left[p(x,y)\right] & =\mathbb{E}\left[- \ln p(x, y)\right] \\
& =-\mathbb{E}\left[\ln p(x,y)\right] \\
& =-\mathbb{E}\left[\ln\left(p(x) p(y)\right)\right] \\
& =-\mathbb{E}\left[\ln p(x) +\ln p(y) \right] \\
& =-\mathbb{E}\left[\ln p(x)\right]-E\left[\ln p(y)\right] \\
& =\operatorname{H}\left[p(x)\right]+\operatorname{H}\left[p(y)\right]
\end{align}

\subsection{Proof of C-BALD}
\label{sec:proof_cbald}
Here, we derive the C-bald objective from Equation \ref{eq:cbald}.
%We use the derivations to create the acquisition function.
%The main idea is to use Assumption 1 to argue that $p(y|x)$ and $p(\ell |x)$, are independent and then use the derivations above to compute the mutual information
\begin{equation}
\begin{aligned}
\mathcal{C}\text{-BALD} &= \operatorname{I}\left[y_i,l, \theta \mid \vec{x}_i\right] \\
&= \mathbb{E}_{p(\theta)}[\operatorname{H}[p_\theta(y_i, \ell_i| \vec{x}_i)] - \operatorname{H}[p(y_i, \ell_i | \vec{x}_i, \theta)]]\, \\ 
&= \mathbb{E}_{p(\theta)}[\operatorname{H}[p_\theta(y_i| \ell_i, \vec{x}_i)]+\operatorname{H}[p_\theta(\ell_i| \vec{x}_i)] - \operatorname{H}[p(y_i | \ell_i, \vec{x}_i, \theta)]-\operatorname{H}[p(\ell_i | \vec{x}_i, \theta)]]\, \\ 
&= \mathbb{E}_{p(\theta)}[\operatorname{H}[p_\theta(y_i|\ell_i, \vec{x}_i)]- \operatorname{H}[p(y_i | \ell_i, \vec{x}_i, \theta)]+\operatorname{H}[p_\theta(\ell_i| \vec{x}_i)]- \operatorname{H}[p_\theta(\ell_i | \vec{x}_i, \theta)]]\, \\ 
&= \underbrace{\mathbb{E}_{p(\theta)}[\operatorname{H}[p_\theta(y_i| \ell_i, \vec{x}_i)]- \operatorname{H}[p(y_i |\ell_i, \vec{x}_i, \theta)]]}_{\text{Mutual information between $y_i$ and $\theta$, assuming we know $\ell$}}+\underbrace{\mathbb{E}_{p(\theta)}[\operatorname{H}[p_\theta(\ell_i| \vec{x}_i)]- \operatorname{H}[p_\theta(\ell_i | \vec{x}_i, \theta)]]}_{\text{Mutual information between $\ell_i$ and $\theta$}}\, \\ 
& = \operatorname{I}\left[y_i, \theta | \ell_i, \vec{x}_i\right] +\operatorname{I}\left[\ell_i, \theta | \vec{x}_i\right] \,.
\end{aligned}
\end{equation}

\section{Additional information of Experiments}
\subsection{Computational resources}
\label{sec:resources}
The \textbf{real} and \textbf{synthetic} data experiments were run parallel on single-core CPUs with varying computing power. The longest experiments ran for 8 hours for the 25 repetitions, i.e. an approximate time to run repetitions in 20 minutes. 
All models are implemented in PyTorch \citep{pytorch}
The experiments on \textbf{SurvMNIST} were run on a GV100 Volta (Tesla V100 - SXM2) with 32GB.
The running time for each scoring function is approximately 25 minutes per 1 experiment.
The 25 repetitions resulted in a total of 12 hours.

\subsection{Synthetic 1D dataset}
\label{sec:synth_1d_app}
The synthetic dataset is generated using a simple sine function with the following censorship,
\begin{align}
x_i &= \mathcal{N}(5,1) \\
y_i^* &= \frac{1}{2}\sin(2x_i) + 2 + \varepsilon_i\\
z_i &= \frac{1}{2}\cos(2x_i) + 2 + \varepsilon_i\\
y_i &= \begin{cases}
    y^*_i & \text{ if } y^*_i \leq z_i \\
    z_i & \text{ else} 
\end{cases} 
\end{align}

and $\ell_i =1 \{y^*_i <= z_i\}$ and $\varepsilon_i\sim\mathcal{N}(0, 0.01| x_i|)$.
We construct the test set using the same approach. However, we extend it as $x\sim U(1.5, 8.5)$. This allows us to evaluate the fit across the entire $x$ range.

\subsection{Real datasets}
\label{sec:real_datasets}
Here, we provide a brief description of the real-world datasets.
Four (GSBG, IHC4, Support, Whas) of the datasets are obtained from \href{https://github.com/jaredleekatzman/DeepSurv/tree/master/experiments/data}{\color{blue}https://github.com/jaredleekatzman/DeepSurv/tree/master/experiments/data}. \citet{Katzman2018deepsurv} provides detail introduction of these datasets. BreastMSK are obtained from \href{https://github.com/TeaPearce/Censored\_Quantile_Regression_NN/tree/main/02_datasets}{\color{blue}https://github.com/TeaPearce/Censored\_Quantile\_Regression\_NN/tree/main/02\_datasets}. \citet{pearce2022censored} provides an introduction to this. 
The churn and credit risk datasets are from \href{https://square.github.io/pysurvival/}{\color{blue} https://square.github.io/pysurvival/}. \citet{pysurvival_cite} provides an introduction to these.
SurvMNIST was introduced in \citep{goldstein2022xcal}.

Here we provide a short introduction to the datasets:

\begin{itemize}
    \item Rotterdam \& German Breast Cancer Study Group (\textbf{GBSG})  requires prediction of survival time for breast cancer patients \citep{GSBG1, GSBG2}. We follow the same pre-processing steps as \citet{Katzman2018deepsurv} and \citet{pearce2022censored}
    \item Molecular Taxonomy of Breast Cancer International
Consortium \textbf{(METABRIC (IHC4))} requires
prediction of survival time for breast cancer patients. Features include clinical and expressions for four genes \citep{Katzman2018deepsurv, pearce2022censored}.
    \item Study to Understand Prognoses Preferences Outcomes
and Risks of Treatment (\textbf{Support}) requires predicting survival time in seriously ill hospitalised patients. The 14 features are age, sex, race, number of comorbidities, presence of diabetes, presence of dementia, presence of cancer, mean arterial blood pressure, heart rate, respiration rate, temperature, white blood cell count, serum sodium, and serum creatinine \citep{SUPPORT}.
    \item Worcester Heart Attack Study (\textbf{WHAS}) requires prediction of acute myocardial infarction
survival. The five features are age, sex, body-mass-index, left heart failure complications and order of MI \citep{lemeshow2011applied, Katzman2018deepsurv, pearce2022censored}
    \item \textbf{BreastMSK} requires prediction of survival time for patients with breast cancer using tumour information. Features include ER, HER, HR, mutation count, and TMB \citep{pearce2022censored}. Original from \href{https://www.cbioportal.org/study/clinicalData?id=breast_msk_2018}{\color{blue} https://www.cbioportal.org/study/clinicalData?id=breast\_msk\_2018}
    \item \textbf{Credit risk}, The task is to predict the time it takes a borrower to repay a loan \citep{pysurvival_cite}. Original from: \href{https://square.github.io/pysurvival/tutorials/credit_risk.html}{\color{blue}https://square.github.io/pysurvival/tutorials/credit\_risk.html}
    \item \textbf{Churn}  is the percentage of customers that stop using a company's products or services. The task is to predict when it will happen \citep{pysurvival_cite}. Original from: \href{https://square.github.io/pysurvival/tutorials/churn.html}{\color{blue} https://square.github.io/pysurvival/tutorials/churn.html}
    \item SurivalMNIST (\textbf{SurvMNIST}) was used in \citet{goldstein2022xcal}, who modified it from Sebastian Pölsterl's blog: \href{https://k-d-w.org/blog/2019/07/survival-analysis-for-deep-learning/}{\color{blue} https://k-d-w.org/blog/2019/07/survival-analysis-for-deep-learning/}. We follow the outline introduced by \citet{pearce2022censored}. 
    It is based on MNIST \href{http://yann.lecun.com/exdb/mnist/}{\color{blue} http://yann.lecun.com/exdb/mnist/}, but each target is drawn from a Gamma distribution according to the class with means [11.25, 2.25, 5.25, 5.0, 4.75, 8.0, 2.0, 11.0, 1.75, 10.75] and variance [0.1, 0.5, 0.1, 0.2, 0.2, 0.2, 0.3, 0.1, 0.4, 0.6].
\end{itemize}

For all the datasets, we log-transform the response $y \sim \log y$.

\subsection{SurvMNIST}
\label{sec:survmnist_archi}
Here we provide a brief overview of the parameters used for the gamma distributions and the model architecture for the SurvMNIST experiments.

\subsection{Parameters for gamma distributions}
We use the same parameters for the gamma distributions as \citet{pearce2022censored}
\begin{table}[!h]
%\vskip 0.15in
\begin{center}
\begin{small}
\begin{sc}
\begin{tabular}{l|rrrrrrrrrrr}
\toprule
\thead{Digit} & \thead{0} & \thead{1} & \thead{2} & \thead{3} & \thead{4} & \thead{5} & \thead{6} & \thead{7} & \thead{8} & \thead{9}
\\ \midrule
Risk & 11.25 & 2.25& 5.25& 5.0& 4.75& 8.0& 2.0& 11.0& 1.75& 10.75\\
Variance&  0.1& 0.5& 0.1& 0.2& 0.2& 0.2& 0.3& 0.1& 0.4& 0.6\\
\bottomrule
\end{tabular}
\caption{Overview of the SurvMNIST dataset and the corresponding gamma distributions and their digits.}
\label{tab:surv_mnist_params}
\end{sc}
\end{small}
\end{center}
\vskip -0.1in
\end{table}

\subsection{Architecture}
The architecture for the SurvMNIST experiments is inspired by the architectures proposed by \citet{pearce2022censored} and \citet{goldstein2022xcal}.

\begin{table}[!h]
\caption{Overview of the SurvMNIST model architecture..}
\label{tab:arch}
%\vskip 0.15in
\begin{center}
\begin{small}
\begin{sc}
\begin{tabular}{ll}
\toprule
& \thead{Layers}
\\ \midrule
& Conv2d(64, ($5\times5$)) \\
& GeLU() \citep{hendrycks2016gaussian} \\
& Consistent Dropout ($0.25$) \citep{gal16dropout} \\
& AvgPool($2\times2$) \\
& Conv2d(128, ($5\times5$)) \\
& GeLU() \\
& Consistent Dropout($0.25$)  \\
& AvgPool ($2\times2$) \\
& Conv2d (258, ($3\times3$)) \\
& GeLU() \\
& Flatten() \\
& Linear (128) \\
& GeLU() \\
& Consistent Dropout($0.25$)\\
\midrule
& Output \\
\bottomrule
\end{tabular}
\end{sc}
\end{small}
\end{center}
\vskip -0.1in
\end{table}

\section{Experiment with different sizes of Bayesian Neural networks}
\label{sec:size_experiments}
Here we provide additional results with varying amount of hidden param

\begin{table}[!h]
\label{tab:sample-table_1}
\vskip 0.15in
\begin{center}
\begin{small}
%\begin{sc}
\resizebox{\textwidth}{!}{
\begin{tabular}{lrr|rrrr}
\toprule
Dataset & Hidden size & Layers & \srandom & \sunc & \sbald & \scbald \\
\midrule
BreastMSK & 64 & 2 & $0.00 \pm 0.00$ & $3.57 \pm 0.67$ &  $0.80 \pm 0.36$ & $-12.49 \pm  0.33$ \\
BreastMSK & 128 & 2 & $0.00 \pm 0.00$ & $9.31 \pm 0.96$ &  $0.74 \pm 0.41$ & $0.62 \pm  0.60$ \\
BreastMSK & 256 & 2 & $0.00 \pm 0.00$ & $6.33 \pm 0.86$ &  $-0.73 \pm 0.36$ & $9.67 \pm  0.81$ \\
BreastMSK & 64 & 3 & $0.00 \pm 0.00$ & $8.29 \pm 1.41$ &  $0.04 \pm 0.73$ & $-12.02 \pm  0.73$ \\
BreastMSK & 128 & 3 & $0.00 \pm 0.00$ & $8.21 \pm 1.43$ &  $-1.89 \pm 0.66$ & $8.75 \pm  1.42$ \\
BreastMSK & 256 & 3 & $0.00 \pm 0.00$ & $5.50 \pm 1.19$ &  $-0.79 \pm 0.68$ & $13.86 \pm  1.55$ \\
BreastMSK & 64 & 4 & $0.00 \pm 0.00$ & $1.44 \pm 1.36$ &  $-0.98 \pm 0.90$ & $-2.18 \pm  1.45$ \\
BreastMSK & 128 & 4 & $0.00 \pm 0.00$ & $5.44 \pm 1.39$ &  $-0.53 \pm 0.78$ & $15.50 \pm  2.01$ \\
BreastMSK & 256 & 4 & $0.00 \pm 0.00$ & $5.05 \pm 1.47$ &  $-0.40 \pm 0.91$ & $21.08 \pm  2.45$ \\
\midrule
metabric & 64 & 2 & $0.00 \pm 0.00$ & $-1.00 \pm 0.39$ &  $1.79 \pm 0.36$ & $29.27 \pm  0.99$ \\
metabric & 128 & 2 & $0.00 \pm 0.00$ & $-0.86 \pm 0.34$ &  $3.86 \pm 0.34$ & $29.12 \pm  1.18$ \\
metabric & 256 & 2 & $0.00 \pm 0.00$ & $-1.48 \pm 0.32$ &  $-0.17 \pm 0.25$ & $33.61 \pm  1.36$ \\
metabric & 64 & 3 & $0.00 \pm 0.00$ & $-1.42 \pm 0.36$ &  $1.84 \pm 0.31$ & $19.20 \pm  0.82$ \\
metabric & 128 & 3 & $0.00 \pm 0.00$ & $-0.67 \pm 0.39$ &  $2.25 \pm 0.34$ & $18.26 \pm  0.94$ \\
metabric & 256 & 3 & $0.00 \pm 0.00$ & $0.15 \pm 0.37$ &  $0.14 \pm 0.30$ & $19.25 \pm  0.76$ \\
metabric & 64 & 4 & $0.00 \pm 0.00$ & $-1.79 \pm 0.36$ &  $-1.69 \pm 0.34$ & $7.42 \pm  0.72$ \\
metabric & 128 & 4 & $0.00 \pm 0.00$ & $-1.47 \pm 0.40$ &  $-1.88 \pm 0.37$ & $7.46 \pm  0.72$ \\
metabric & 256 & 4 & $0.00 \pm 0.00$ & $1.78 \pm 0.51$ &  $0.04 \pm 0.41$ & $15.47 \pm  0.93$ \\
\midrule
whas & 64 & 2 & $0.00 \pm 0.00$ & $0.46 \pm 0.24$ &  $1.22 \pm 0.18$ & $-11.63 \pm  0.17$ \\
whas & 128 & 2 & $0.00 \pm 0.00$ & $0.37 \pm 0.20$ &  $0.32 \pm 0.12$ & $-4.28 \pm  0.17$ \\
whas & 256 & 2 & $0.00 \pm 0.00$ & $1.86 \pm 0.20$ &  $0.83 \pm 0.10$ & $2.22 \pm  0.16$ \\
whas & 64 & 3 & $0.00 \pm 0.00$ & $-0.01 \pm 0.36$ &  $0.68 \pm 0.23$ & $-5.81 \pm  0.40$ \\
whas & 128 & 3 & $0.00 \pm 0.00$ & $0.42 \pm 0.27$ &  $1.68 \pm 0.17$ & $0.26 \pm  0.32$ \\
whas & 256 & 3 & $0.00 \pm 0.00$ & $-2.89 \pm 0.25$ &  $-1.92 \pm 0.15$ & $2.84 \pm  0.27$ \\
whas & 64 & 4 & $0.00 \pm 0.00$ & $3.19 \pm 0.36$ &  $0.35 \pm 0.26$ & $-2.22 \pm  0.43$ \\
whas & 128 & 4 & $0.00 \pm 0.00$ & $-0.40 \pm 0.30$ &  $-0.33 \pm 0.19$ & $2.89 \pm  0.36$ \\
whas & 256 & 4 & $0.00 \pm 0.00$ & $-1.09 \pm 0.24$ &  $-1.02 \pm 0.19$ & $6.46 \pm  0.35$ \\
\midrule
gsbg & 64 & 2 & $0.00 \pm 0.00$ & $1.51 \pm 0.07$ &  $0.20 \pm 0.06$ & $5.85 \pm  0.06$ \\
gsbg & 128 & 2 & $0.00 \pm 0.00$ & $0.28 \pm 0.07$ &  $1.70 \pm 0.05$ & $6.85 \pm  0.08$ \\
gsbg & 256 & 2 & $0.00 \pm 0.00$ & $0.20 \pm 0.06$ &  $0.81 \pm 0.05$ & $7.31 \pm  0.08$ \\
gsbg & 64 & 3 & $0.00 \pm 0.00$ & $-0.07 \pm 0.04$ &  $-0.11 \pm 0.04$ & $3.86 \pm  0.04$ \\
gsbg & 128 & 3 & $0.00 \pm 0.00$ & $-0.81 \pm 0.05$ &  $-0.04 \pm 0.05$ & $5.58 \pm  0.05$ \\
gsbg & 256 & 3 & $0.00 \pm 0.00$ & $1.20 \pm 0.06$ &  $-1.20 \pm 0.05$ & $5.67 \pm  0.07$ \\
gsbg & 64 & 4 & $0.00 \pm 0.00$ & $0.12 \pm 0.04$ &  $0.06 \pm 0.03$ & $1.49 \pm  0.04$ \\
gsbg & 128 & 4 & $0.00 \pm 0.00$ & $0.90 \pm 0.05$ &  $0.81 \pm 0.03$ & $3.00 \pm  0.04$ \\
gsbg & 256 & 4 & $0.00 \pm 0.00$ & $1.36 \pm 0.05$ &  $0.28 \pm 0.04$ & $5.08 \pm  0.06$ \\
\midrule
support & 64 & 2 & $0.00 \pm 0.00$ & $0.80 \pm 0.02$ &  $-0.08 \pm 0.02$ & $5.81 \pm  0.02$ \\
support & 128 & 2 & $0.00 \pm 0.00$ & $0.07 \pm 0.02$ &  $-0.27 \pm 0.01$ & $5.98 \pm  0.02$ \\
support & 256 & 2 & $0.00 \pm 0.00$ & $0.53 \pm 0.02$ &  $-0.71 \pm 0.01$ & $5.83 \pm  0.02$ \\
support & 64 & 3 & $0.00 \pm 0.00$ & $0.57 \pm 0.02$ &  $-0.42 \pm 0.01$ & $4.56 \pm  0.02$ \\
support & 128 & 3 & $0.00 \pm 0.00$ & $0.70 \pm 0.02$ &  $-0.53 \pm 0.01$ & $4.55 \pm  0.02$ \\
support & 256 & 3 & $0.00 \pm 0.00$ & $-0.03 \pm 0.02$ &  $-0.44 \pm 0.01$ & $5.17 \pm  0.02$ \\
support & 64 & 4 & $0.00 \pm 0.00$ & $-0.43 \pm 0.02$ &  $-0.46 \pm 0.01$ & $2.67 \pm  0.01$ \\
support & 128 & 4 & $0.00 \pm 0.00$ & $0.35 \pm 0.02$ &  $-0.79 \pm 0.01$ & $4.04 \pm  0.02$ \\
support & 256 & 4 & $0.00 \pm 0.00$ & $1.39 \pm 0.02$ &  $-0.18 \pm 0.01$ & $4.30 \pm  0.02$ \\
\bottomrule
\end{tabular}
}
%\end{sc}
\end{small}
\end{center}
\caption{Relative decrease in the area under the curve (RD-AUC) compared to the \srandom scoring function. A higher value in the table represents better performance, with the best performance highlighted in \textbf{bold}.}
\vskip -0.1in

\end{table}

\begin{table}[!h]
\label{tab:sample-table_2}
\vskip 0.15in
\begin{center}
\begin{small}
%\begin{sc}
%\resizebox{\textwidth}{!}{
\begin{tabular}{lrr|rrrr}
\toprule
Dataset & Hidden size & Layers & \srandom & \sunc & \sbald & \scbald \\
\midrule
Churn & 64 & 2 & $0.00 \pm 0.00$ & $0.56 \pm 0.19$ &  $0.10 \pm 0.17$ & $20.87 \pm  0.47$ \\
Churn & 128 & 2 & $0.00 \pm 0.00$ & $4.88 \pm 0.25$ &  $1.25 \pm 0.16$ & $18.79 \pm  0.42$ \\
Churn & 256 & 2 & $0.00 \pm 0.00$ & $-0.32 \pm 0.23$ &  $-0.05 \pm 0.11$ & $19.46 \pm  0.50$ \\
Churn & 64 & 3 & $0.00 \pm 0.00$ & $1.49 \pm 0.29$ &  $0.38 \pm 0.21$ & $29.77 \pm  0.74$ \\
Churn & 128 & 3 & $0.00 \pm 0.00$ & $4.10 \pm 0.30$ &  $0.08 \pm 0.20$ & $32.29 \pm  0.81$ \\
Churn & 256 & 3 & $0.00 \pm 0.00$ & $5.40 \pm 0.32$ &  $0.24 \pm 0.17$ & $30.57 \pm  0.92$ \\
Churn & 64 & 4 & $0.00 \pm 0.00$ & $-0.66 \pm 0.23$ &  $0.66 \pm 0.19$ & $29.17 \pm  0.49$ \\
Churn & 128 & 4 & $0.00 \pm 0.00$ & $3.88 \pm 0.26$ &  $0.06 \pm 0.16$ & $33.34 \pm  0.64$ \\
Churn & 256 & 4 & $0.00 \pm 0.00$ & $2.12 \pm 0.20$ &  $0.30 \pm 0.13$ & $35.96 \pm  0.85$ \\
\midrule
Credit risk & 64 & 2 & $0.00 \pm 0.00$ & $-0.48 \pm 0.29$ &  $-0.40 \pm 0.29$ & $15.37 \pm  0.45$ \\
Credit risk & 128 & 2 & $0.00 \pm 0.00$ & $-1.08 \pm 0.28$ &  $-0.15 \pm 0.27$ & $16.87 \pm  0.46$ \\
Credit risk & 256 & 2 & $0.00 \pm 0.00$ & $1.00 \pm 0.25$ &  $0.03 \pm 0.21$ & $21.53 \pm  0.45$ \\
Credit risk & 64 & 3 & $0.00 \pm 0.00$ & $-0.57 \pm 0.32$ &  $-0.47 \pm 0.30$ & $20.21 \pm  0.47$ \\
Credit risk & 128 & 3 & $0.00 \pm 0.00$ & $0.50 \pm 0.32$ &  $0.71 \pm 0.28$ & $21.17 \pm  0.51$ \\
Credit risk & 256 & 3 & $0.00 \pm 0.00$ & $0.99 \pm 0.33$ &  $0.16 \pm 0.30$ & $23.53 \pm  0.61$ \\
Credit risk & 64 & 4 & $0.00 \pm 0.00$ & $-0.48 \pm 0.29$ &  $1.26 \pm 0.28$ & $17.60 \pm  0.42$ \\
Credit risk & 128 & 4 & $0.00 \pm 0.00$ & $0.07 \pm 0.28$ &  $-0.27 \pm 0.28$ & $18.38 \pm  0.44$ \\
Credit risk & 256 & 4 & $0.00 \pm 0.00$ & $1.43 \pm 0.28$ &  $-0.21 \pm 0.23$ & $19.40 \pm  0.46$ \\
\bottomrule
\end{tabular}
%}
%\end{sc}
\end{small}
\end{center}
\caption{Relative decrease in the area under the curve (RD-AUC) compared to the \srandom scoring function. A higher value in the table represents better performance, with the best performance highlighted in \textbf{bold}.}
\vskip -0.1in

\end{table}

\end{document}